\PassOptionsToPackage{numbers}{natbib}
 
 \documentclass{article}
 \pdfoutput=1

     \usepackage[preprint]{neurips_2019}

\usepackage[utf8]{inputenc} 
\usepackage[T1]{fontenc}    
\usepackage{hyperref}       
\usepackage{url}            
\usepackage{booktabs}       
\usepackage{amsfonts}       
\usepackage{nicefrac}       
\usepackage{microtype}      

\usepackage{amsmath, amssymb, amsthm, mathtools, bm}
\usepackage{geometry}
\usepackage{graphicx, xcolor}
\usepackage{algorithm, algpseudocode}
\usepackage{todonotes}

\newcommand{\argmax}[1]{\underset{#1}{\operatorname{argmax}}\;}

\newcommand{\be}{\begin{equation}}
\newcommand{\ee}{\end{equation}}
\newcommand{\ba}{\begin{align}}
\newcommand{\ea}{\end{align}}
\newcommand{\bea}{\begin{eqnarray}}
\newcommand{\eea}{\end{eqnarray}}

\usepackage{amsmath}

\title{Inverse Rational Control with Partially Observable Continuous Nonlinear Dynamics}

\author{
    Saurabh Daptardar \\
    Department of ECE \\
    Rice University \\
    Houston, TX 77005 \\
    \texttt{svd3@rice.edu} \\
    \And
    Paul Schrater \\
    Department of Computer Science \\
    University of Minnesota \\
    Minnesota, IN 12345 \\
    \texttt{schrater@umn.edu} \\
    \And
    Xaq Pitkow \\
    Department of Neuroscience \\
    Baylor College of Medicine \\
    Houston, TX 77005 \\
    \texttt{xaq@rice.edu} \\
}

\begin{document}
\let\b\mathbf

\newcommand{\Q}{Q^{\pi}}
\newcommand{\Qo}{Q^{*}}
\newcommand{\V}{V^{\pi}}
\newcommand{\Vo}{V^{*}}
\newcommand{\R}{\mathbb{R}}
\newcommand{\D}{\mathcal{D}}
\newcommand{\bPhi}{\mathbf{\Phi}}
\newcommand{\Beta}{\bm{\beta}}
\newcommand{\T}{\top}
\newcommand{\I}{\mathcal{I}}
\newcommand{\N}{\mathcal{N}}

\newcommand{\Rew}{R_{\theta_r}}
\newcommand{\rew}{r_{\theta_r}}
\newcommand{\w}{w^{\pi}}

\def\NoNumber#1{{\def\alglinenumber##1{}\State #1}\addtocounter{ALG@line}{-1}}

\theoremstyle{definition}
\newtheorem{definition}{Definition}[section]

\maketitle

\begin{abstract}
Continuous control and planning remains a major challenge in robotics and machine learning. Neuroscience offers the possibility of learning from animal brains that implement highly successful controllers, but it is unclear how to relate an animal's behavior to control principles. Animals may not always act optimally from the perspective of an external observer, but may still act rationally: we hypothesize that animals choose actions with highest expected future subjective value according to their own internal model of the world. Their actions thus result from solving a different optimal control problem from those on which they are evaluated in neuroscience experiments. With this assumption, we propose a novel framework of model-based inverse rational control that learns the agent's internal model that best explains their actions in a task described as a partially observable Markov decision process (POMDP). In this approach we first learn optimal policies generalized over the entire model space of dynamics and subjective rewards, using an extended Kalman filter to represent the belief space, a neural network in the actor-critic framework to optimize the policy, and a simplified basis for the parameter space. We then compute the model that maximizes the likelihood of the experimentally observable data comprising the agent's sensory observations and chosen actions. Our proposed method is able to recover the true model of simulated agents within theoretical error bounds given by limited data. We illustrate this method by applying it to a complex naturalistic task currently used in neuroscience experiments. This approach provides a foundation for interpreting the behavioral and neural dynamics of highly adapted controllers in animal brains.
\end{abstract}

\section{Introduction}

Brains evolved to understand, interpret, and act upon the physical world. To thrive and reproduce in a harsh and dynamic natural environment, animals therefore evolved flexible, robust controllers. Machine learning and neuroscience both aim to emulate or understand how these successful controllers operate.

Traditional imitation learning or inverse control methods extract reusable policies to help generate autonomous behavior or help predict future actions from an agent's past behavior. In contrast, here we do not aim to extract the policy of a well-trained expert. Instead we want to identify the internal model and preferences of a real agent that may make interesting mistakes \citep{lakshminarasimhan2018dynamic}. Unlike the convention in robotics and artificial intelligence, animals are not optimized for stable, narrowly-defined tasks, but instead survive by performing well enough in competitive, changing ecological niches. Comparing actual control policies across conditions, animals, or species may ultimately guide us to broad control principles that generalize beyond specific tasks. Meanwhile, our approach estimates latent assumptions and dynamic beliefs for real biological controllers, thereby providing targets for understanding neural network representations and implementations of perception and action.

Our approach does not assume that agents perform optimally at a given task. Rather, we assume that agents are {\it rational} --- by which we mean that agents act optimally according to their own internal model of the task, which may differ from the actual task.

We solve this problem by formulating the agent's policy as an optimal solution of a Partially Observable Markov Decision Process (POMDP) \citep{sutton1998reinforcement} that the agent assumes that it faces. Whereas Reinforcement Learning (RL) tries to find the optimal policy given the dynamics and the reward function, Inverse Reinforcement Learning (IRL, \citep{russell1998learning,choi2011inverse,babes2011apprenticeship}) tries to find a reward function which best accounts for observed actions from an agent. Similarly, inverse optimal control (IOC) \citep{dvijotham2010inverse,schmitt2017see} tries to find the assumed model dynamics that explains observed actions with a given cost function yields the similar optimal policy. Trying to find both is, in general, an ill-posed problem, but with a sufficiently constrained model or proper conditioning it can be solved. Others have solved this problem in the fully observable setting \citep{herman2016inverse,Reddy/etal/18b}.

Some of the present authors recently solved the partially observable case for a discrete state space with discretized beliefs \citep{wu2018inverse}. However, while that solution was useful for their particular application, in general its computational expense grows rapidly with the problem complexity and size, and that solution assumes that actions are easy to select from the discretized value function. These issues makes that solution infeasible for continuous state spaces and continuous controls. In the present work, we provide methods that address these challenges.

Two major questions are how to construct representations of the agent's beliefs, and how to choose a policy based on those beliefs. We solve the first by assuming that agents know the basic task structure but not the task parameters, so we can use model-based inference to identify the agent's beliefs. We solve the second by using flexible function approximation to estimate values and policies over the parameter space. We demonstrate the utility of our approach by applying it to simulated agents in a continuous control task with latent variables that has been used in neuroscience: catching fireflies in virtual reality \citep{lakshminarasimhan2018dynamic}.

\section{Prior work on Inverse Reinforcement Learning}

Inverse Reinforcement Learning (IRL) and Inverse Optimal Control (IOC) solve aspects of the general problem of inferring internal models of an observed agent. Ng and Russell \citep{ng2000algorithms} formulate the IRL problem as a linear programming problem which takes agents policy as input and learn reward parameters with an $\ell_2$ regularization which makes the agent's policy optimal and also maximizes difference between the value at optimal action and value at next best action for each state (maximizing the margin of optimal policy). Abbeel and Ng \citep{abbeel2004apprenticeship} extend the idea to perform maximum margin IRL. 
Max margin IRL introduces certain biases to solve the ambiguity of multiple reward functions. Some methods \citep{ziebart2008maximum} resolve the ambiguity by adopting a principle of maximum entropy \citep{jaynes1957information} to obtain a distribution over reward functions without bias. Max Entropy IRL \citep{ziebart2008maximum} tries to estimate a distribution over trajectories which has maximum entropy under the constraints of matching expectations of certain features. 
This approach is feasible only for finite discrete states and actions.
and becomes intractable as the length of trajectories increases.

For IOC, the Relative Entropy IRL framework \citep{boularias2011relative} allows for unknown dynamics, and minimizes the KL divergence between two distributions over trajectories.
The analytical solution needs a transition dynamics function which is estimated by importance sampling. It is important to note here that they do not assume the agent behaves optimally in any sense. This method was extended in Guided Cost Learning by \citep{finn2016guided} to model a free maximum entropy formulation where the reward function is represented by a neural network. They use reinforcement learning to move the baseline policy towards policies with higher rewards imposing weak optimality.

A more principled way to think about this problem class is to view the state action trajectories as observations about a reward and dynamics with latent parameters \citep{wu2018inverse}. \citep{ramachandran2007bayesian} define the likelihood of trajectories as a exponential Boltzmann distribution in state action value function $Q(s, a; r)$. Their inference method performs a random walk over the reward parameter space using the posterior, and compute the posterior over the parameters by computing the optimal action value function $Q^*$. This is a nested loop approach that is feasible only for small spaces, as computing the optimal policy and value functions in the inner loop is costly. Other notable approaches include maximum likelihood IRL \citep{vroman2014maximum,babes2011apprenticeship}, Path Integral IRL \citep{kalakrishnan2013learning}, simultaneous estimation of rewards and fully-observed dynamics \citep{herman2016inverse}. \citep{kalakrishnan2013learning} proposes learning a continuous state, continuous action reward function by sampling locally around the optimal agent trajectories. These methods are local searches that do not consider trajectories with unexpected control input.

Across all of these methods, there is not a complete inverse solution that can learn how an agent models rewards, dynamics, and uncertainty in a partially observable task with continuous nonlinear dynamics and continuous controls.
\section{Inverse Rational Control}

To define the Inverse Rational Control problem, we first formalize our tasks as Partially Observable Markov Decision Processes (POMDPs). A POMDP $M$ is a tuple, $M=(S,A,\Omega,R,T,O,\gamma)$ that includes states $s\in S$, actions $a\in A$, observations $o\in\Omega$, reward or loss function $R$, as well as transition probabilities $T(s'|s,a)$, observation probabilities $O(o|s)$, and a temporal discount factor $\gamma$. Since the states $s$ are only partially observable, the POMDP determines the agent's time-dependent `belief' about the world, namely a posterior over states given the history of observations and actions, $b_t=p(s_t|o_{1:t},a_{1:t})$. An optimal solution of the POMDP determines a value function $Q(b,a)$ through the Bellman equation \citep{bellman1957dynamic}, which in turn defines a policy $\pi$ that selects the action that maximizes the value. We parameterize $R$, $T$ and $O$, and therefore the value function and policy, by $\theta=(\theta_r,\theta_t,\theta_o)$.

We define the Inverse Rational Control problem as identifying, using only data measurable from an agent's behavior, the most probable internal model parameters $\theta\in\Theta$ for an agent that solves the POMDP described above. Specifically, we allow measurements of the states and action trajectory taken by the agent, but we have no access to the beliefs that motivate those actions. The agent's sensory observations may be fully observed, fully unobserved, or partially observed, for instance when we cannot access the particular observation noise that corrupts the agent's senses. A solution to IRC provides both parameters and an estimate of the agent's beliefs over time.

A core idea in our approach is first to learn policies and value functions over the parameterized manifold of models, reflecting an optimized {\it ensemble} of agents, rather than a single optimized agent. This then allows us to maximize the likelihood of the observed data generated by an unknown agent, by finding which parameters from the ensemble best explain the agent's actions. Because our overarching goal is scientific, a major benefit of this approach is to provide the best interpretable explanation of an agent's behaviors as rational for a task as defined by some parameters $\theta$.

\subsection{General formulation of optimal control ensembles}

There are two difficulties in solving the POMDP model ensemble and computing a likelihood function over trajectories. First, policies and value functions are complex functions of the model parameters. We address this by using flexible, trainable networks to learn the value function and policies, as described below. Second, the belief updating process of the agent involves a difficult integral and requires correct handling of the agent's observations.

The agent does not observe the world states $s$, but gets partial observations $o$ about them that induce a belief $b_t=p(s_t|o_{1:t},a_{1:t},\theta)$. When the agent plans into the future, it does not know the future observations, so it must marginalize over them to predict the consequences of its actions $a$. If an agent were given the observations, future beliefs would be known, but when observations are unknown the agent has only a distribution over beliefs arising from an average transition probability $\bar{T}_{\theta_d}$ between belief states, $\bar{T}_{\theta_d}(b'|b,a)=\int do\, T_{\theta_t}(b'|b,a,o)O_{\theta_o}(o|a,b)$. The parameters of this average transition probability subsume the parameters from the transitions and observation functions, $\theta_d=(\theta_t,\theta_o)$. Once we have formulated the problem completely in terms of belief states, and identified a tractable finite representation of them, then we can use many of the tools developed for fully observed Markov Decision Processes (MDP).

An agent chooses its actions based on these beliefs according to a policy $\pi$ as $a\sim\pi(a|b)$. An optimal policy is the one that produces the maximal (temporally discounted) expected future reward, which can be computed using the Bellman equation,
\begin{align}
\Q_{\theta}(b,a)&= \Rew(b,a) + \gamma \iiint
da' db' do'\  T_{\theta_t}(b'| b, a) O_{\theta_o}(o|b,a)\pi(a' | b') \Q_{\theta}(b', a') \\
&= \Rew(b,a) + \gamma \iint
da' db' \  \bar{T}_{\theta_d}(b' | b, a) \pi(a' | b') \Q_{\theta}(b',a')\label{eq:belief_lspi_cont}
\end{align}

In practice we replace the full posterior $b$ by a parametric form (such as a multivariate Gaussian) for which the transitions can be computed tractably, although this may introduce approximation errors.

To solve the inverse problem by inferring an agent's internal model that guides their actions, we begin with the simple observation that the optimal policy $\pi^*(a | b;\theta)$ and optimal state-action value function $Q^*(b,a;\theta)$ function change depending on the dynamics and the rewards of the problem. That is, the optimal policy and state-action value function should implicitly be functions of the dynamics and reward parameters. We will make this explicit by assuming that the dynamics and the reward function are parameterized functions denoted by $\bar{T}_{\theta_d}(b' | b, a )$ and $\Rew(b, a)$. We choose this parametric form wisely to impose and exploit structure in the task.

Although the parametric form for the dynamics and reward may be simple or convenient, the resultant dependence of $\Q$ on these parameters is usually more complicated. Therefore we represent $\Q$ by a flexible function class. This class could be a family of end-to-end trainable deep networks, for example, or a sum of basis functions as used in nonlinear support vector machines. We may learn the $Q_\theta$ using any of a variety of reinforcement learning methods. In our work we trained a function approximator using rewards obtained by policies across many parameters $\theta$ sampled from a uniform prior on parameter space $\Theta$, although a good prior over the parameters could make learning faster and more robust.

The optimal policy is to pick the best actions according to the updated value function. The rewards obtained by these actions can then provide additional information about the value function, so we may repeat the above procedure iteratively until the value function converges. However, even choosing the best action from a given $\Q$ can be difficult. When there are few allowed actions, then the policy is easy to implement: one simply exhaustively evaluates $\Q$ for all actions and picks the most valuable. When there is a continuum of actions, such that this approach is infeasible, then must choose another method to maximize value. Here we select actions by approximating the policy using an actor-critic method implemented by a supplemental network that is trained to optimize the value function by Deep Deterministic Policy Gradient (DDPG) \citep{lillicrap2015continuous}, although we could also use other versions of Policy Iteration or $Q$-learning.

Our algorithm for Policy Iteration across model space is given below:
\begin{algorithm}[H]
\caption{Learning optimal value functions across parameter space} \label{algo:our1}
\begin{algorithmic}[1]
    \Statex Given a class of POMDP problems over parameter space $\Theta$
    \State Initialize a random policy $\pi$
    \Repeat
        \State Sample multiple $\theta \sim \mathcal{U}(\Theta)$ and store in $\D_{\theta}$
        \ForAll{$\theta \in \D_{\theta}$}
            \State Generate $(b, a)$ trajectories $\tau$ using policy $\pi_\theta$ and store in $\mathcal{T}$
        \EndFor
        \State Solve equation \ref{eq:belief_lspi_cont} for $Q^{\pi}$
        \State Improve policy $\pi$ using $Q^{\pi}$ by either $\epsilon$-greedy updates, softmax action selection \citep{sutton1998reinforcement}, or policy gradient for continuous actions. 
    \Until policy $\pi$ converges
    \State \Return $\pi^* \gets \pi$
\end{algorithmic}
\end{algorithm}

\subsection{Inverse Rational Control from optimal control ensembles}

Now that we have constructed an ensemble of approximately optimal policies over the task space, we now aim to recover the true parameters of an agent whose behavior we observe. Specifically, we try to find parameters which maximize the likelihood of the agent's trajectories. For a given $\theta \in \Theta$, we know the transition dynamics and have already computed the optimal policy.

Given the observable state and action trajectories indexed by $i\in\{1\ldots N\}$, the log likelihood of the model parameters $\theta$ given the experimental observations of one trajectory $i$ is
\begin{align}
    \label{eq:likelihood}
    \mathcal{L}_i=\log{p(s^i_{0:T}, a^i_{0:T}| o^i_{0:T}, \theta)} = \log p(s^i_0) + \sum_t \log\pi(a^i_t | b^i_t(o^i_{0:t}), \theta) + \log p(s^i_{t+1} | a^i_t, s^i_t, \theta)
\end{align}
Note that this likelihood is conditioned on observations $o$, which the agent has but the external observer does not. When we do not have the observations but only state action sequence and the initial state, ideally we would marginalize the likelihood in equation \ref{eq:likelihood} over all latent observation trajectories, like in traditional Expectation Maximization \citep{dempster1977maximum}. Here instead we use a hard E step, taking the maximum {\it a posteriori} (MAP) value for the observations given the observed state, $\hat{o}={\rm argmax}_{o} O_{\theta_o}(o|s)$. To maximize this likelihood, we rely on auto-differentiation to compute the gradient of the policy.
The algorithm for learning the maximum likelihood estimate of true parameters given a state-action trajectory is given by Algorithm \ref{algo:our2}.
\begin{algorithm}[htb]
\caption{Maximum Likelihood estimate for agent's parameters} \label{algo:our2}
\begin{algorithmic}[1]
    \State Initialize $\theta$ randomly by sampling from the prior $\theta \sim \mathcal{P}(\Theta)$
    \Repeat
        \State Estimate agent's observations by MAP inference, $\hat{o}=\argmax{o} O_{\theta_o}(o|s)$.
        \State Estimate beliefs $b$ given $\hat{o}$ using $\hat{b}_t=p(s_t|\hat{o}_{1:t},a_{1:t})$.
        \State Compute total log-likelihood over trajectories
        \begin{equation*}
        \mathcal{L} \gets \sum\nolimits_i \mathcal{L}_i
        \end{equation*}
        where $\mathcal{L}_i$ is computed according to equation \ref{eq:likelihood}
        \State Update $\theta$ by one gradient ascent step with learning rate $\alpha$
        \begin{equation*}
            \theta \gets \theta + \alpha \nabla_{\theta} \mathcal{L}
        \end{equation*}
    \Until $\mathcal{L}$ converges
    \State \Return $\theta$
\end{algorithmic}
\end{algorithm}

In Algorithm \ref{algo:our1} we estimate optimal policies over the entire parameter space. When we then estimate the particular $\theta$ that best explains the agent's actions, it is by construction consistent with rational behavior: the best fit policy is optimal given those parameters, even if they do not match the actual task. Of course, for real data, this model fit does not imply that the agent's policy actually falls into our model class. Where needed, the model class should be expanded to accommodate other potentially false assumptions the agent may make.

We can summarize the whole Inverse Rational Control framework in Algorithm \ref{algo:ourcomb}:
\begin{algorithm}[htb]
\caption{Parameter space Inverse Rational Control} \label{algo:ourcomb}
\begin{algorithmic}[1]
    \State Define a family of POMDPs for the task, parametrized by $\Theta$.
    \State Learn the state-action value functions $Q(b,a;\theta)$ over $\theta\in\Theta$ using Algorithm \ref{algo:our1}.
    \State Estimate agent's parameters $\hat{\theta}$ from its observable behavior using Algorithm \ref{algo:our2}.
\end{algorithmic}
\end{algorithm}

\section{Demonstration task: `Catching fireflies'} \label{sec:application}

We demonstrate that our proposed Inverse Rational Control framework works by recovering the internal model of simulated agents performing two control tasks. Ultimately we will apply this approach to understand the internal control models of behaving animals in neuroscience experiments, where we do not know the ground truth. However, using simulated agents allows us to verify the method when we know the ground truth.

\begin{figure}[htb]
    \begin{center}
        \includegraphics[width=\linewidth]{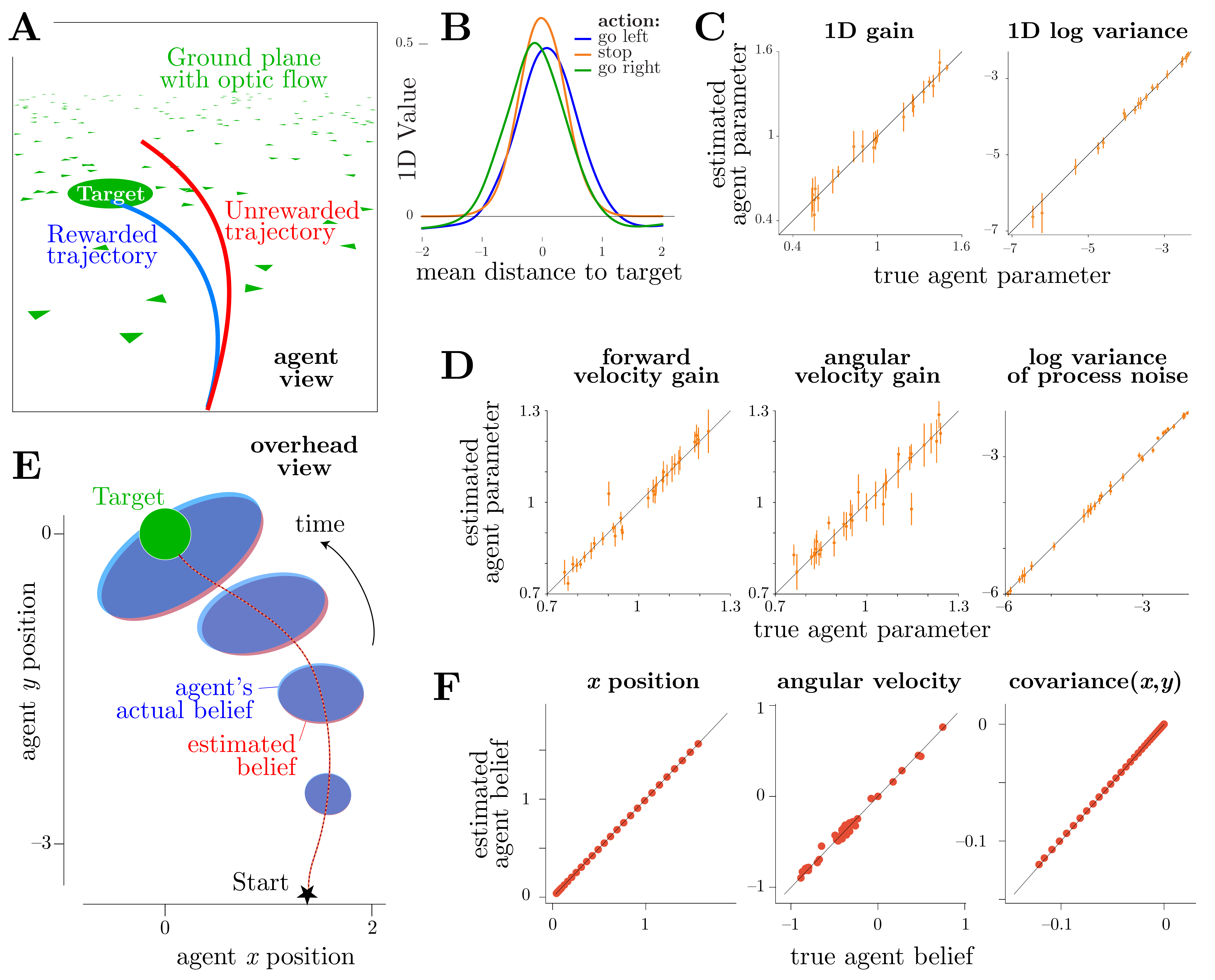}
    \end{center}
    \caption{Firefly control task \citep{lakshminarasimhan2018dynamic}. ({\bf A}) To reach the transiently visible firefly target, an agent must navigate by noisy optic flow over a dynamic textured plane. ({\bf B}) For a 1D version of this task with only three allowed actions $a$, we derive sensible state-action value functions $Q(b,a;\theta)$, here showing that it is best to move toward the target and then stop, unless the target is too far to justify the effort. Our method accurately recovers the agent's assumed parameters within limits imposed by the data, both for the 1D task ({\bf C}) and the 2D task ({\bf D}), as shown for several example parameters $\theta$ inferred from different agents. Error bars show 95\% confidence intervals derived from the curvature of the likelihood given limited data. ({\bf E}) Overhead view illustrates one agent's belief dynamics, depicted by posterior covariance ellipses centered at each believed location (blue), as well as our method's estimates of those beliefs (red). ({\bf F}) The estimated and true belief dynamics closely match. Three representative components of the belief representation are shown here: most likely firefly $x$ position, most likely angular velocity, and the posterior covariance between $x$ and $y$.}
    \label{fig:application}
\end{figure}

In this task, an agent must navigate through a virtual world to catch a flashing target, called the `firefly' (Figure \ref{fig:application}A) \citep{lakshminarasimhan2018dynamic}. When the agent stops moving, the trial ends, the agent receives a reward if it is sufficiently close to the firefly position, and a new target appears. Each target is visible only briefly, and sensory inputs provide partial (noisy) observations of self motion, so the agent is uncertain about the current position of its target as well as its current velocity.

The two example problems both use this task structure, either in one or two dimensions. In the 1D task, the world state and control variables are therefore location and forward velocity, and we only allow three discrete control outputs. In the 2D task we add angle and angular velocity to the state space, allow continuous control outputs, and allow the agent to make noisy observations of its state. Given the uncertainty in its state estimate, the agent must track its belief over the states, and plan its optimal trajectory based on this belief state space and unknown future belief transitions based on unknown future observations. Given this task setting, the agent maximizes its reward expected over the distribution implied by the belief. Each episode or trial is a variable duration $T$ depending on when the agent stops moving. The goal is to maximize the total rewards over this finite time horizon.

Our first simulation experiment is a simplified version of the task, where the agent's position is restricted to one dimension, state dynamics are linear, and the agent receives no observations. The belief state is Gaussian-distributed over the agent's relative location to the target, with scalar mean $\mu$ and variance $\sigma^2$ evolving as
\begin{align}
    \mu_{t+1} &= \mu_t + g_a a \hspace{1cm}
    \sigma^2_{t+1} = \sigma^2_t + \sigma^2_0 
\end{align}
The model parameters are the action gain $g_a$ and the process noise log standard deviation $\log{\sigma_0}$. We have only three discrete actions, namely: go left, go right, and stop.

To compute the optimal policies we use a $Q$-learning variant of Algorithm \ref{algo:our1}.
Recall that we can substitute different learning algorithms, but it is crucial to learn the optimal action value function $Q^*$ or policy as a function of $\theta$. One approach we took for the 1D demonstration task was to construct $Q$ from a factorized set of basis functions. We experimented with both manually constructed polynomial and exponential basis functions, and a collection of random shallow networks as basis functions for both $\phi(b,a)$ and $\psi(\theta)$. We found that the random network basis was more expressive and performed better for this task. 
Figure \ref{fig:application}B plots the value function $Q^*(s, a, \theta)$ for each action and a fixed $\theta$.

Now that we have computed the optimal policies over a manifold of parameters, we create an agent with one particular set of parameters $\theta^*$. Using the corresponding optimal policy and belief dynamics, we simulate this agent's belief state and action trajectories. Next we use experimentally observable data (the action sequence) to update the belief trajectory per our model. Given this data we follow the maximum likelihood estimation as described in Algorithm \ref{algo:our2} using the most probable belief states to estimate the parameters. In Figure \ref{fig:application}C we plot the estimated parameters recovered by our algorithm against the agent's true parameters, along with the 95\% confidence interval for parameter $\theta_i$ as given by $2(\I^{-1/2})_{ii}$ where $\I$ is the Fisher information matrix. In almost all cases the true value is within the resultant error bars, showing that we correctly recover the true parameters within the precision allowed by the data. 

Our second simulation uses the full 2D firefly task. We use two gain parameters, one on forward velocity input and another on angular velocity input, and one process noise parameter that affects $x$ and $y$ position identically, with no angular process noise. We assume the agent maintains a Gaussian representation for its beliefs over the world state, which it updates using an extended Kalman filter given the dynamics parameters $\theta_d$.

To compute the optimal policies we use Deep Deterministic Policy Gradient (DDPG) \citep{lillicrap2015continuous} in Algorithm \ref{algo:our1}. We sample the parameters uniformly over a moderate range and for each $\theta$, we run the current policy $\pi_\theta$ to collect sequences of belief states, actions, and rewards. Every time point $(\hat{b}_t,a_t,r_t)$ in these sequences along with the corresponding $\theta$ was provided as input to two neural networks, each three layers deep, 64 units wide, and using softplus activations, whose output estimated the state-action value function and the policy. The learning updates for this are same as the DDPG algorithm.

As before, we create an agent by choosing parameters $\theta^*$, and generate trajectories using corresponding belief dynamics and optimal policy. We preserve only the experimentally measurable action trajectories, and apply Algorithm \ref{algo:our2} to compute the maximum likelihood estimate and confidence intervals. Figure \ref{fig:application}D again shows that we can recover agents' true parameters up to the intrinsic uncertainty. Additionally, we compare our estimates of their beliefs to their actual beliefs and find excellent agreement (Figure \ref{fig:application}E,F).

\section{Summary}
To summarize our proposed inverse reinforcement learning framework, we express and learn the optimal action value function generalized over the task parameter space. This can be thought of as learning optimal value function and policy for all the parameters in the parameter space. For large, complex tasks, using an informative prior with sampling focused on relevant regions of parameter space could greatly accelerate the learning and make the approximation more robust. 

Most other related frameworks have a nested inner loop of policy optimization or refinement to adapt to the optimal policy for the new updated parameter. Unlike such methods, our method separates these optimizations into two separate loops, one for learning optimal policy over parameter space, and second for maximum likelihood estimation of the true parameters given the optimal policy computed in the previous loop.

\subsection{Limitations}
Firstly, our assumption parametrized dynamics and reward function induces obvious model bias. Of course we can reduce this bias by making the parameter space richer and more flexible by increasing the number and variety of parameters. However, this would come with a price in computability and interpretability.

In case of POMDPs, where we need to work with posterior distribution over the states or the belief, deriving analytical parametric belief update equations can be a difficult inference problem in general. For special cases like finite discrete states, the posterior updates can be succinctly represented matrix products, parameterized by transition matrices. For continuous state and Gaussian noises, we can use Kalman filter updates which are parameterized by the gain and covariance matrices, and for small nonlinearity this can be generalized using extended Kalman filters. But for general cases with general posterior distributions, a belief update might be a difficult inference problem. Though this is not necessarily a limitation of our approach but a challenge for inference problems more generally, still it may be difficult to apply our method in such cases.

Similarly, as stated several times above, the learning in Algorithm \ref{algo:our1} can be replaced with any suitable reinforcement learning algorithm which guarantees convergence to an optimal policy, as long as the policy or the $Q$ function is explicitly a function of model parameters. Our framework theoretically does not limit the number of parameters as long as there is some algorithm which can explore and learn optimal policy over that large space.

\subsection{Outlook}

One interesting novelty of our framework is to make optimal action value $Q^*$ and optimal policy $\pi^*$  explicit functions of the parameters $\theta$. We can use this representation to extend the control task to hierarchical models: Instead of thinking of the parameters as fixed in the environment, we could consider them to be slowly changing latent variables describing task demands in a dynamic world. The value function over this larger space would then provide a means of adaptive control, where the agent could adjust its policy based on its current belief about $\theta$. To use IRC for such an agent, we would then need to introduce higher-level parameters that describe the full dynamics.Now, this looks like any other optimal control problem, and when we solve for optimal control problem for $\tilde{M}$ as a function of augmented state $\tilde{s}$, we are learning the optimal control for $M_{\theta}$ generalized over $\theta$.

We have implemented Inverse Rational Control for neuroscience applications, but the core principles have value in other fields as well. We can view IRC as a form of Theory of Mind, whereby one agent (a neuroscientist) creates a model of another agent's mind (for a behaving animal). Theory of Mind is a prominent component of social interactions, and imputing rational motivations to actions provides a useful description of how people think \citep{baker2011bayesian}. Designing useful artificial agents to interact with others in online environments would also benefit from being able to attribute rational strategies to others. One important example is self-driving cars, which currently struggle to handle the perceived unpredictability of humans. While humans do indeed behave unpredictably, some of this may stem from ignorance of the rational computation that drives actions. Inverse Rational Control provides a framework for better interpretation of other agents, and serves as a valuable tool for greater understanding of unifying principles of control.

\subsubsection*{Acknowledgments}

The authors thank Kaushik Lakshminarasimhan, Dora Angelaki, Greg DeAngelis, James Bridgewater and Minhae Kwon for useful discussions. SD and XP were supported in part by the Simons Collaboration on the Global Brain award 324143. PS and XP were supported in part by BRAIN Initiative grant NIH 5U01NS094368. XP was supported in part by an award from the McNair Foundation, NSF CAREER Award IOS-1552868, and NSF 1450923 BRAIN 43092-N1.

\bibliographystyle{plain}
\bibliography{references.bib}{}

\end{document}